\pdfoutput=1

\documentclass[11pt]{article}

\usepackage[final]{naacl2021}
\usepackage{times}
\usepackage{latexsym}
\usepackage[T1]{fontenc}
\usepackage[utf8]{inputenc}
\usepackage{microtype}
\usepackage{graphicx}
\usepackage{amsmath}
\usepackage{amssymb}
\usepackage{booktabs}
\usepackage{enumitem}
\usepackage{amsbsy}
\usepackage{array}
\usepackage{cleveref}

\newcommand{\footnotehref}[2]{\footnote{\href{#1}{#2}}}

\newcommand{\CSEN}{CS$\rightarrow$EN}
\newcommand{\ENCS}{EN$\rightarrow$CS}
\newcommand{\ENDE}{EN$\rightarrow$DE}

\renewcommand{\arraystretch}{1.25}
\newcommand{\staroff}{\hspace*{0.16cm}$\star$\hspace*{-0.32cm}}

\graphicspath{{img/}}

\title{Sampling and Filtering of Neural Machine Translation Distillation Data}

\author{Vilém Zouhar \\
  Institute of Formal and Applied Linguistics, Charles University \\
  \texttt{zouhar@ufal.mff.cuni.cz}}

\begin{document}
\maketitle
\begin{abstract}
In most of neural machine translation distillation or stealing scenarios, the goal is to preserve the performance of the target model (teacher).
The highest-scoring hypothesis of the teacher model is commonly used to train a new model (student). If reference translations are also available, then better hypotheses (with respect to the references) can be upsampled and poor hypotheses either removed or undersampled. 

This paper explores the importance sampling method landscape (pruning, hypothesis upsampling and undersampling, deduplication and their combination) with English to Czech and English to German MT models using standard MT evaluation metrics.
We show that careful upsampling and combination with the original data leads to better performance when compared to training only on the original or synthesized data or their direct combination.
\end{abstract}

\graphicspath{{img/}}
\section{Introduction}

Model distillation is a process of transferring the knowledge of one or more, usually larger, model(s) into another, usually smaller, model \cite{bucilua2006model}. A variation of this is training a new model in a way that its performance is similar to that of the already trained one. This is achieved by making use of either teacher predictions (black-box) or other products of the workings of the teacher, such as attention-score or decoder score (grey/glass-box).
Assuming we have access to a parallel corpus, we focus on sampling the translation hypotheses and making use not only of the teacher scores but also of their comparison to the reference.

There are various possible motivations for model distillation. The student model can be much smaller than the teacher model, which has the benefit of faster inference speed \cite{germann-EtAl:2020:WMT}. It can also be used for model stealing, where an adversary tries to copy the teacher functionality. This is a practical concern for production-level MT systems \cite{wallace2020imitation}.

One of the approaches for knowledge distillation is to use the teacher model to generate a new dataset for the student model to train on. Having access to a trained teacher model, this approach does not require parallel data and can leverage large monolingual corpora. Reference translations, however, help with determining which of the teacher's translations are good and which are of low quality. 

We focus on this approach and propose and compare several importance sampling approaches to prepare training data for student models that leverage access to reference translations. These include pruning, upsampling and undersampling, deduplication and their combination. We show that a combination of these methods improves the student performance over just using the reference or the best hypothesis (by the decoder score), which is a common distillation practice.

The experiment code is available open-source.\footnotehref{https://github.com/zouharvi/reference-mt-distill}{github.com/zouharvi/reference-mt-distill}

\subsection{Related work}

The general methodology for knowledge distillation in the form of teacher-student has been proposed by \citet{hinton2015distilling}. 
For the MT task specifically, \citet{tan2019multilingual} focus on vastly reducing the number of parameters, while retaining the performance of a multi-lingual teacher. \citet{wei2019online} and \citet{gordon2020distill} use distillation during training to further improve the model performance.

The work of \citet{kim2016sequence} shows that taking either the top sentence with respect to the teacher decoder score or BLEU \cite{metrics_bleu} improves the performance.
\citet{germann-EtAl:2020:WMT} presented student models that distil knowledge from a larger teacher model with a negligible loss in performance. They manipulate the queried data based on target sentence quality, such as by removing sentences that are not correctly recognized by a language identifier. For the parallel part of the data, they extract the best BLEU scoring sentence out of 8 hypotheses. 
\citet{freitag2017ensemble} experiment with pruning sentences that are below some TER \cite{metrics_ter} threshold (lower is better). They further document the effect of using an ensemble of teachers and also reducing the student model size. 

\section{Methods}

The evaluation of every sampling method follows the following three-step process. First, the specific parallel corpus (\Cref{subsec:data}) is translated by the teacher model (\Cref{subsec:models}) for the considered translation direction. New datasets based on metrics are then created. The reference is taken into consideration during the hypothesis selection. We train new models (students) on these datasets and measure their performance. 
There are 12 hypotheses (default in Marian NMT) provided by the teacher using beam search for every source sentence which we consider when composing a new dataset.

\begin{figure}[h!]
    \center
    \includegraphics[width=0.6\linewidth]{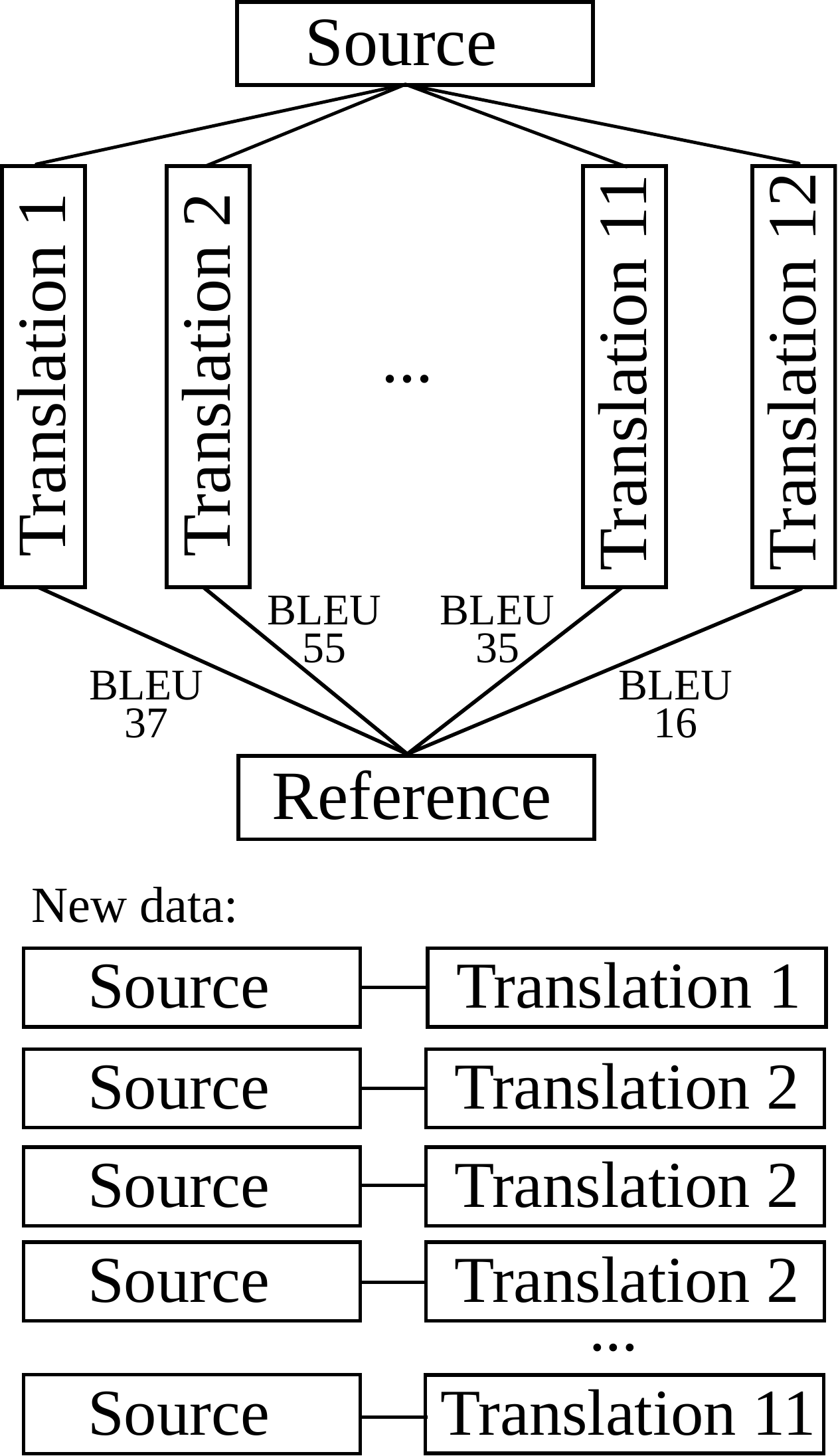}
    \vspace{0.15cm}
    \caption{Scheme of an example of hypothesis sampling with BLEU metric.}
    \label{fig:intro_scheme}
    \vspace{-0.2cm}
\end{figure}

\Cref{fig:intro_scheme} shows an example of the sampling process with BLEU. Twelve translations are made of \textit{Source} and each receives a score against the provided reference. The new data contain \textit{Translation 2} three times, because of its high score. \textit{Translation 12} is omitted because of its low score. This upsampling is explained in detail in \Cref{subsec:sampling}.

\subsection{Data} \label{subsec:data}

We make use of the Europarl v10 parallel corpus \cite{koehn2005europarl} for English-Czech (0.6M sentences) and English-German (1.8M sentences). The sentences are longer (23 target words per sentence on average) than in the WMT News Task domain \cite{barrault2020findings}. 
To modern standards, this dataset is relatively small and very domain restricted. This was chosen deliberately because of computational limitations.\footnote{$\sim$4500 GPU hours in total for the whole experiment} Despite that it demonstrates the results of the different sampling methods with respect to each other. These results may not be transferable to large parallel corpora in which training data is abundant.

For every language pair, we randomly sample 15k sentences as development dataset (used only for determining the best epoch and early stopping) and 15k sentences for final test evaluation which is reported. The WMT News test dataset is not used for student evaluation, because the students are trained on a limited amount of data and on a different domain. Out of the WMT20 News tokens, $0.18\%$ are not present in the Europarl training set. This would introduce a higher variance into the WMT News test evaluation, which would be largely dependent on the diversity of the teacher vocabulary.

\subsection{Models} \label{subsec:models}

The teachers\footnote{Version \textit{student.base} at \href{https://github.com/browsermt/students}{github.com/browsermt/students}} in this experiment are transformer-based \cite{vaswani2017transformer}, speed optimized and were themselves created by knowledge distillation from state-of-the-art models \cite{popel2020transforming, junczys2019microsoft}, as proposed by \citet{germann-EtAl:2020:WMT}. The Czech$\leftrightarrow$English model is described by \citet{germann-EtAl:2020:WMT} and the English$\rightarrow$German model by \citet{model_deen}. Our student models follow the teacher's architecture with half the size of the embedding vector (256 instead of 512) and half of the attention heads (4 instead of 8). Student models were trained with an early stopping of 20 evaluations on validation data with evaluation performed every 10k sentences. Vocabularies were not shared from the teacher because they did not affect the results, and not using them makes fewer assumptions regarding the level of access to the teacher model. 
Marian NMT \cite{mariannmt} is used for teacher decoding and student training.

\Cref{tab:teacher_performance} shows the teacher performance measured on WMT20 News and the test subset of Europarl. Czech models performed better on the Europarl than on the News task, while for the German model the trend was the opposite. This may be caused by the fact that the models were distilled from a system that had Europarl as part of the training data, CzEng 2.0 \cite{kocmi2020announcing}.

\begin{table}[h!]
    \renewcommand{\arraystretch}{1.0}
    \center
    \begin{tabular}{lccc}
        \toprule
        Dataset: & \CSEN & \ENCS & \ENDE \\
        \midrule
        BLEU: & & & \\
        WMT20 News\hspace{-0.5cm} & $28.2$ & $35.8$ & $42.7$ \\
        Europarl & $46.1$ & $38.2$ & $32.1$ \\
        \midrule
        ChrF: & & & \\
        WMT20 News\hspace{-0.5cm} & $0.57$ & $0.55$ & $0.66$ \\
        Europarl & $0.69$ & $0.64$ & $0.61$ \\
        \midrule
        TER: & & & \\
        WMT20 News\hspace{-0.5cm} & $0.57$ & $0.71$ & $0.51$ \\
        Europarl & $0.41$ & $0.50$ & $0.61$ \\
        \bottomrule
    \end{tabular}
    \caption{Teacher models BLEU, ChrF and TER scores on WMT20 News Task dataset and Europarl domain.}
    \label{tab:teacher_performance}
    \vspace{-0.3cm}
\end{table}

\subsection{Sampling} \label{subsec:sampling}

Concerning the sampling metrics (always between the considered hypothesis and the reference), we make use of BLEU, ChrF \cite{metrics_ChrF}, TER (negative), the difference (negative of absolute value) in subword unit counts by SentencePiece \cite{sentencepiece} ($\text{SP}$) and decoder probability divided by the number of output tokens ($\text{score}$). TER and SP are negative in \Cref{sec:results} so that higher is always better. The motivation for SP is to capture the difference in length of the hypotheses with respect to the reference. This is a very naive metric, but we can use it to see the performance and the behaviour of all the other metrics. Although BLEU is a document-level metric, it can also be used to determine sentence similarity. Standard machine translation metrics\footnote{
    Sacrebleu metrics version strings:\\
    BLEU+case.mixed+numrefs.1+smooth.exp+tok.13a+v1.4.14\\
    ChrF2+numchars.6+space.false+v1.4.14\\
    TER+tok.tercom-nonorm-punct-noasian-uncased+v1.4.14
} are computed using Sacrebleu \cite{sacrebleu}. Different sampling methods are used even though the goal is to maximize the BLEU scores of the student models. There is no reason to assume that sampling only based on BLEU will lead to the best results.

The number of training sentences differs for every method. We define the following notation.

\begin{itemize}[itemsep=3pt,leftmargin=17pt]
    \item $T$ - top; $T_\text{metric}^n$ takes $n$ top translation hypotheses according to \textit{metric}; equal to $S_\text{metric}^{1, 1, \ldots 1 (n)}$. The student model may benefit from seeing e.g. the second best hypothesis, even though it's not the best available.
    This results in $n$ times the number of original sentences which are all different.
    
    \item $S$ - skewed; $S_\text{metric}^{k_1, k_2, \ldots k_n}$ takes $k_1\times$ the top translation hypotheses according to \textit{metric}, $k_2\times$ the second top translation, etc. As opposed to $T_\text{metric}^n$, this method tries to preserve the information of the ordering by setting $k_1 \ge k_2 \ge \ldots k_n$.
    This results in $(\sum k_i)$ times the number of original sentences but only $n$ times of which are different sentences.
    
    \item $Dedup[X]$ deduplicates sentence pairs of $X$. It is used after joining the results of other methods. This method is useful for emulating the \textit{or} operation: $Dedup[A + B]$ then means ``all sentences in either $A$ or $B$.'' The output size is strictly dependent on their overlap. 
    
    \item $G$ - greater than; $G_\text{metric}^{m}$ takes all sentence translations with \textit{metric} at least $m$. This results in sentences that are close to the reference according to the metric.
    The number of output sentences highly dependent on the threshold and is discussed in the corresponding section.
\end{itemize}

Sampling methods can be combined: $T_\text{bleu}^2 + G_\text{score}^{-10}$ joins the top 2 sentences measured by BLEU and adds them to the hypotheses with decoder score of at least $-10$. Duplicates are intentionally not removed; thus, hypotheses in both sampling methods are upsampled.

\section{Results} \label{sec:results}

\paragraph{Baseline.} \Cref{tab:b_results} shows results for baseline sampling methods. \textit{Original} corresponds to training only to the provided parallel corpus (references). $T^1_\text{score}$ takes only the highest-scoring hypothesis from the decoder, which is related to the scenario where the reference is not available, and the decoder score is the best measure for hypothesis quality.\footnote{MT quality estimation tools could be used to approximate the sentence translation quality or language models to use sentence fluency in lieu of translation quality.} The sampling method $T^{12}_{-}$ takes all available hypotheses (metric does not matter).

\begin{table}[h!]
    \center
    \begin{tabular}{lccc}
        \toprule
        Dataset & \CSEN & \ENCS & \ENDE \\
        \midrule
        Original & $41.6$ & $31.8$ & $25.1$ \\
        $T^1_\text{score}$& $40.0$ & $31.2$ & $28.5$ \\
        $T^{12}_{-}$ & $41.1$ & $31.6$ & $28.4$ \\
        \bottomrule
    \end{tabular}
    \caption{BLEU scores for students trained on baseline datasets}
    \label{tab:b_results}
\end{table}

Training on the original data leads to better results than training on the best scoring hypotheses. Training on all hypotheses results in slightly lower BLEU performance. This may be caused by the small amount of training data available in which case taking all hypotheses just improves the vocabulary and language modelling capacity.

\paragraph{Best hypotheses.} The results of datasets created by taking either the best one or the four best hypotheses for every source sentence is shown in \Cref{tab:mn_results}. In the case of multiple hypotheses having the same score, the one with the highest decoder score is chosen. The top one and top four hypotheses were chosen to show that the optimum is neither the top one nor the top twelve (all) hypotheses.

On average, the hypothesis overlap\footnote{Overlap computed as $\text{average}_{m1 \ne m2} {|T^1_{m1} \cap T^1_{m2}|}/{n}$ and $\text{average}_{m1 \ne m2} {|T^4_{m1} \cap T^4_{m2}|}/{(4 n)}$. Original data size is $n$.} in sampling between metrics is $29\%$ for $T^1$ and $51\%$ for $T^4$. This is expected and shows that when more top hypotheses are taken into the new dataset, the individual metrics tend to matter less.

\begin{table}[h!]
    \center
    \begin{tabular}{lccc}
        \toprule
        Dataset & \CSEN & \ENCS & \ENDE \\
        \midrule
        $T^1_\text{BLEU}$ & $42.6$ & $34.4$ & $29.5$ \\
        $T^1_\text{ChrF}$ & $43.8$ & $33.9$ & $30.5$ \\
        $T^1_\text{TER}$ & $43.0$ & $36.1$ & $28.5$ \\
        $T^1_\text{SP}$ & $39.9$ & $29.5$ & $28.2$ \\
        \midrule
        $T^4_\text{BLEU}$ & $44.0$ & $33.3$ & $29.3$ \\
        $T^4_\text{ChrF}$ & $44.3$ & $34.9$ & $29.6$ \\
        $T^4_\text{TER}$ & $44.2$ & $32.0$ & $28.8$ \\
        $T^4_\text{SP}$ & $41.8$ & $32.3$ & $27.9$ \\
        $T^4_\text{score}$ & $44.2$ & $32.0$ & $28.8$ \\
        \bottomrule
    \end{tabular}
    \caption{BLEU scores for students trained on best-one and best-four hypotheses datasets}
    \label{tab:mn_results}
\end{table}

Taking only the top-scoring hypothesis of reference-based metrics, $T^1$ showed better results than the baseline (training on the original data, taking the highest decoder scoring hypothesis or taking all hypotheses).
In all cases the $T^4$ outperformed $T^1$.
The main gains were on \CSEN{} and \ENCS{}. Although the results on \ENDE{} are only slightly better than the baseline, they are systematic across all metrics except for SP.
The effect of choosing the metric for the top four hypotheses seems marginal, even compared to sampling based on the decoder score. The only exception is the SP difference, which leads to lower results.

\paragraph{Thresholding.}

Determining a single threshold for all datasets leads to a vastly different number of hypotheses being selected (the use of $G^{65}_\text{BLEU}$ results in $1.3\times$ the original dataset for \CSEN, but $0.6$ for \ENDE). Therefore, we establish different metric thresholds for every dataset so that the new datasets are $1\times$ to $1.5\times$ the original size for consistent results across language pairs.

Some of the source sentences were easier to translate, and more of their hypotheses were put into the new dataset. Others had no hypothesis above a given threshold and were not included in the new data at all. On average only $25\%$ of original sentences were preserved for BLEU, ChrF, TER and SP. For the decoder score metric, it is $46\%$. The high loss of source sentences is expected since most of the hypotheses share large portions of the target sentence and only differ in a few words. All of them will then behave similarly with respect to the metric.

\begin{table}[ht!]
    \vspace{0.15cm}
    \center
    \begin{tabular}{llll}
        \toprule
        \multicolumn{1}{c}{Dataset} & 
        \multicolumn{1}{c}{\CSEN} & 
        \multicolumn{1}{c}{\ENCS} & 
        \multicolumn{1}{c}{\ENDE} \\
        \midrule
        $G_\text{BLEU}$ &
        $39.0_{\,65}$ &
        $30.2_{\,60}$ &
        $27.2_{\,55}$ \\
        $G_\text{ChrF}$ &
        $37.4_{\,0.82}$ &
        $29.2_{\,0.81}$ &
        $26.5 _{\,0.80}$ \\
        $G_\text{TER}$ &
        $37.8_{\,-0.2}$ &
        $30.2_{\,-0.25}$ &
        $25.2_{\,-0.24}$ \\
        $G_\text{SP}$ &
        $32.5_{\,\text{--}1}$ &
        $19.6_{\,\text{--}2}$ &
        $23.0 _{\,\text{--}1}$ \\
        $G_\text{score}$ &
        $39.0_{\,\text{--}0.08}$ &
        $32.0_{\,\text{--}0.09}$ &
        $27.6_{\,\text{--}0.11}$ \\
        \bottomrule
    \end{tabular}
    \caption{BLEU scores for students trained on datasets made of hypotheses above threshold of different metrics. Metrics thresholds are in subscript.}
    \label{tab:g_results}
\end{table}

The highest performance is achieved using $G_\text{score}$ which can be explained by how much of the original sentences were preserved.
$G_\text{score}$ shows that it is possible to achieve a performance comparable to $T^1_\text{score}$ with less than half of the source sentences by only taking all hypotheses with a decoder score above a threshold. $G_\text{BLEU}$ gets worse results (on average $-1.1$ BLEU), but with only $27\%$ source sentences preserved.

Better performance could be achieved by lowering the threshold to allow more source sentences and by intersecting the result with some of the other sampling methods, thus eliminating only the very low-quality sentence pairs. This is the approach (done with 5 hypotheses) done by \citet{freitag2017ensemble}: $T^1_{score} \cap G^{-0.8}_{TER}$.

\paragraph{Upsampling.}

In the first upsampling case, $S^{4,3,2,1}$, the best hypothesis is present four times, the second-best three times, the third-best two times and the fourth-best once. The reason for upsampling better hypotheses is that we want to force the optimizer to make bigger steps for sentence pairs that are of high quality, but at the same time, we want to present other hypotheses to enlarge the vocabulary and improve the student's language model. The most straightforward approach is to put multiple copies of the high-quality example into the dataset. We also experiment with $S^{2,2,1,1}$, because the upsampling intensity for every hypothesis rank is an independent variable as well. Both of these schemes are relatively conservative so that they can be compared to each other and to $T^4$. Results for upsampling within a single metric are shown in \Cref{tab:s_results}.

\begin{table}[ht!]
    \vspace{0.15cm}
    \center
    \begin{tabular}{lccc}
        \toprule
        Dataset & \CSEN & \ENCS & \ENDE \\
        \midrule
        $S^{4,3,2,1}_\text{BLEU}$ & $\boldsymbol{45.2}$ & $\boldsymbol{37.1}$ & $29.7$ \\
        $S^{4,3,2,1}_\text{ChrF}$ & $42.9$ & $36.6$ & $\boldsymbol{30.1}$ \\
        $S^{4,3,2,1}_\text{TER}$ & $44.4$ & $36.9$ & $29.8$ \\
        $S^{4,3,2,1}_\text{SP}$ & $41.8$ & $30.7$ & $28.5$ \\
        $S^{4,3,2,1}_\text{score}$ & $41.4$ & $33.7$ & $27.9$ \\
        \midrule
        $S^{2,2,1,1}_\text{BLEU}$ & $44.3$ & $36.5$ & $29.6$ \\
        $S^{2,2,1,1}_\text{ChrF}$ & $45.2$ & $36.1$ & $29.8$ \\
        $S^{2,2,1,1}_\text{TER}$ & $43.5$ & $33.4$ & $29.6$ \\
        $S^{2,2,1,1}_\text{SP}$ & $41.8$ & $33.3$ & $28.9$ \\
        $S^{2,2,1,1}_\text{score}$ & $43.5$ & $33.4$ & $29.6$ \\
        \bottomrule
    \end{tabular}
    \caption{BLEU scores for students trained on datasets made by upsampling top hypotheses within a single metric using $S^{4,3,2,1}$ and $S^{2,2,1,1}$}
    \label{tab:s_results}
\end{table}

Both versions of upsampling ($S^{4,3,2,1}$  and $S^{2,2,1,1}$) outperformed all of the previous results. There seems to be no systematic difference between $S^{4,3,2,1}$ and $S^{2,2,1,1}$. With the exception of SP and decoder score, the metrics are comparable.
A direct comparison can be made to $T^4 = S^{1,1,1,1}$ because both $T^4$ and the upsampling methods contain all source sentences and even the same hypotheses. The only difference is that in the upsampling case, the better hypothesis is upsampled.
In this case $S^{2,2,1,1}$ had higher results over $T^4$ with $p<0.005$ by Student's t-test.\footnote{Average was subtracted from the three directions so that $T^4$ and $S^{2,2,1,1}$ could be treated as only two distributions.}

\paragraph{Combination.}

For the combination scenarios, the newly sampled datasets are joined together. This is shown in \Cref{tab:c_results}. In the first four cases, the sampling methods were joined with the original data. A baseline to this is $T^1_\text{score} + \text{Original}$, which is commonly used for distillation. 

Deduplicating the top four hypotheses according to BLEU or decoder score and adding them to the original data did not improve over the baseline. Combining the upsampling according to the decoder score with the original data also did not help. Replacing the decoder score with BLEU resulted in a significant improvement. The original data is upsampled so that the ratio of synthetic to original data is 4:1 in the first case and 2:1 in the second one.

For the rest of the cases, the methods are combined without the original data. Baselines are shown in \Cref{tab:b_results}. The combination of the top four hypotheses ($T^4_\text{BLEU}$ or $T^4_\text{score}$) with all of the hypotheses, $T^{12}_-$, improved over the baseline, including $T^{12}_-$, but performed poorly with respect to the other methods.
Taking hypotheses that are in the top four according to either BLEU or decoder score leads to the best results in this section. The top one hypothesis, according to BLEU, is upsampled at least two and at most four times. This seems to work best for \ENDE{} where the training data were three times larger.

\begin{table*}[h!]
    \center
    \begin{tabular}{lccc}
        \toprule
        \multicolumn{1}{l}{Dataset} & \CSEN & \ENCS & \ENDE \\
        \midrule
        $T^1_\text{score} + \text{Original}$ & $44.4$ & $36.4$ & $28.3$ \\
        \cmidrule{1-1}
        $Dedup[T^4_\text{BLEU} + T^4_\text{score}] +\text{Original}$ & $43.7$ & $35.3$ & $29.1$ \\
        $S^{4,3,2,1}_\text{score} + 2\times\text{Original}$ & $43.9$ & $36.1$ & $28.3$ \\
        $S^{4,3,2,1}_\text{BLEU} + 2\times\text{Original}$ & $\boldsymbol{45.5}$ & $\boldsymbol{37.3}$ & $28.8$ \\
        $S^{4,3,2,1}_\text{BLEU} + 4\times\text{Original}$ & $\boldsymbol{45.5}$\staroff & $\boldsymbol{37.4}$\staroff & $28.9$ \\
        \cmidrule{1-1}
        $T^4_\text{score} + T^{12}_-$ & $41.6$ & $33.2$ & $28.3$ \\
        $T^4_\text{BLEU} + T^{12}_-$ & $42.6$ & $33.9$ & $28.7$ \\
        $T^4_\text{BLEU} + T^4_\text{score}$ & $43.3$ & $33.2$ & $28.9$ \\
        $Dedup[\sum T^{2}_\text{metric}]$ & $43.6$ & $34.7$ & $29.1$ \\
        $Dedup[\sum T^{2}_\text{metrics}] + T^{12}_-$ & $40.8$ & $32.0$ & $27.2$ \\
        $Dedup[T^4_\text{BLEU} + T^4_\text{score}] + T^1_\text{BLEU} + T^1_\text{score}$ & $43.5$ & $34.7$ & $29.2$ \\
        $Dedup[T^4_\text{BLEU} + T^4_\text{score}] + Dedup[T^1_\text{BLEU} + T^1_\text{score}]$ & $42.6$ & $34.9$ & $\boldsymbol{29.6}$\staroff \\
        $Dedup[T^4_\text{BLEU} + T^4_\text{score}]$ & $43.5$ & $35.0$ & $\boldsymbol{29.3}$  \\
        \bottomrule
    \end{tabular}
    \caption{BLEU scores for students trained on datasets made of combination of sampling methods. $\sum_\text{metric}$ sums over all used metrics (BLEU, ChrF, TER, SP, score).}
    \label{tab:c_results}
\end{table*}

\paragraph{Bigger student model.} 

To demonstrate the data sampling method behaviour on slightly larger models, the common distillation baseline ($T^1_\text{score} + \text{Original}$) and the best performing proposed sampling method ($S^{4,3,2,1}_\text{BLEU} + 4\times\text{Original}$) were used to train a student of the same size as the used teacher (embedding vector dimension 512 and 8 attention heads). The results are shown in \Cref{tab:c_results_big}. They are systematically higher than for the smaller models, and the difference between the baseline and the best sampling is preserved.

\begin{table*}[h!]
    \center
    \begin{tabular}{lccc}
        \toprule
        \multicolumn{1}{l}{Dataset} & \CSEN \hspace{-0.15cm} & \ENCS \hspace{-0.15cm} & \ENDE \\
        \midrule
        $T^1_\text{score} + \text{Original}$ & $44.7$ & $36.2$ & $28.3$ \\
        \cmidrule{1-1}
        $Dedup[T^4_\text{BLEU} + T^4_\text{score}] +$ Original & $44.3$ & $36.2$ & $28.5$ \\
        $S^{4,3,2,1}_\text{BLEU} + 2\times\text{Original}$\hspace{-0.2cm} & $\boldsymbol{46.9}$ & $\boldsymbol{38.5}$ & $\boldsymbol{28.8}$\\
        $S^{4,3,2,1}_\text{BLEU} + 4\times\text{Original}$\hspace{-0.2cm} & $\boldsymbol{47.4}$\staroff & $\boldsymbol{38.9}$\staroff & $\boldsymbol{28.9}$ \\
        \bottomrule
    \end{tabular}
    \caption{BLEU scores for students trained on datasets made of combination of top hypothesis and original data. Trained with parameters equal to the teacher's: embedding vector dimension 512 and 8 attention heads.}
    \label{tab:c_results_big}
    \vspace{-0.3cm}
\end{table*}

\section{Summary}
\vspace{-0.1cm}

Although widely used, taking only the highest-scoring sentence (with respect to the decoder score or any reference-based metrics, such as BLEU) does not lead to the best results. In the context of the proposed experiments, these are achieved by a combination of top hypotheses and the original data, such as $S_\text{BLEU}^{4,3,2,1} + 4\times\text{Original}$ (upsampling according to BLEU and joining with the original data duplicated four times). Here, an improvement of an average $+2$ BLEU points against $T^1_\text{score}$ + Original was achieved.

The choice of the sampling metric does not significantly influence the results, especially in cases where more than the top one hypothesis is sampled. Because of this, in most scenarios the decoder score can be used instead, reducing the need for translation references.

\paragraph{Future work.}

We worked with only two upsampling schemes: $S^{4,3,2,1}$ and $S^{2,2,1,1}$. However, the two vectors are arbitrary and more of the vast vector space should be explored, especially with more than the top four hypotheses considered or more skewed towards the best hypothesis. More sophisticated methods based on the value of the metric instead of just the ordering could also be tried out.

The effects of large models (both teacher and student) and data access should be explored to verify the transferability of the results of the current setup. Specifically, the teacher model should not be a distilled model itself. The robustness of the training should also be established.

Even though this paper focused solely on MT, the importance sampling methods could also be applied and verified on other NLP tasks, possibly even on more general machine learning problems.

\section*{Acknowledgements}

Sincere thanks to Philipp Zimmermann, Martin Popel and both PBML reviewers for their helpful suggestions and comments.

This study was supported by the Czech Science Foundation (grant n. 19-26934X, NEUREM3).

The work described also used services provided by the LINDAT/CLARIAH-CZ Research Infrastructure (\href{https://lindat.cz}{lindat.cz}), supported by the Ministry of Education, Youth and Sports of the Czech Republic (Project No. LM2018101).

\bibliography{anthology,custom}
\bibliographystyle{acl_natbib}

\end{document}